\documentclass[10pt,twocolumn,letterpaper]{article}

\usepackage[pagenumbers]{cvpr} 

\usepackage[dvipsnames]{xcolor}
\usepackage{multirow}
\usepackage{amssymb}

\definecolor{cvprblue}{rgb}{0.21,0.49,0.74}
\usepackage[pagebackref,breaklinks,colorlinks,citecolor=cvprblue]{hyperref}
\usepackage{algorithm}
\usepackage{algpseudocode}
\def\paperID{3396} 
\def\confName{CVPR}
\def\confYear{2024}

\title{Decision Boundary-aware Knowledge Consolidation Generates Better Instance-Incremental Learner }

\author{Qiang Nie$^1$,
Weifu Fu$^2$, Yuhuan Lin$^2$, Jialin Li$^2$, Yifeng Zhou$^2$, Yong Liu$^2$, Lei Zhu $^1$, Chengjie Wang$^2$\\
1. Hong Kong University of Science and Technology (Guangzhou)\\
2. Tencent Youtu Lab\\
{\tt\small \{qiangnie, leizhu\}@hkust-gz.edu.cn}\\
{\tt\small \{ryanwfu,gleelin,jarenli,joefzhou,choasliu,jasoncjwang\}@tencent.com}
}

\begin{document}
\maketitle
\begin{abstract}
Instance-incremental learning (IIL) focuses on learning continually with data of the same classes. Compared to class-incremental learning (CIL), the IIL is seldom explored because IIL suffers less from catastrophic forgetting (CF). However, besides retaining knowledge, in real-world deployment scenarios where the class space is always predefined, continual and cost-effective model promotion with the potential unavailability of previous data is a more essential demand. Therefore, we first define a new and more practical IIL setting as \textbf{promoting the model's performance besides resisting CF with only new observations}. Two issues have to be tackled in the new IIL setting: 1) the notorious catastrophic forgetting because of no access to old data, and 2) broadening the existing decision boundary to new observations because of concept drift. To tackle these problems, our key insight is to moderately broaden the decision boundary to fail cases while retain the old boundary. Hence, we propose a novel decision boundary-aware distillation method with consolidating knowledge to teacher to ease the student learning new knowledge. We also establish the benchmarks on existing datasets Cifar-100 and ImageNet. Notably, extensive experiments demonstrate that the teacher model can be a better incremental learner than the student model, which overturns previous knowledge distillation-based methods treating student as the main role.
\end{abstract}    
\section{Introduction}
\label{sec:intro}
In recent years, many excellent deep-learning-based networks are proposed for variety of tasks, such as image classification, segmentation, and detection. Although these networks perform well on the training data, they inevitably fail on some new data that is not trained in real-world application.
Continually and efficiently promoting a deployed model's performance on these new data is an essential demand. Current solution of retraining the network using all accumulated data has two drawbacks: 1) with the increasing data size, the training cost gets higher each time, for example, more GPUs hours and larger carbon footprint~\cite{patterson2021carbon}, and 2) in some cases the old data is no longer accessible because of the privacy policy or limited budget for data storage. 
In the case where only a little or no old data is available or utilized, retraining the deep learning model with new data always cause the performance degradation on the old data, \ie, the catastrophic forgetting (CF) problem. To address CF problem, incremental learning~\cite{rebuffi2017icarl,he2011incremental,douillard2020podnet,wei2023online}, also known as continual learning, is proposed.
Incremental learning significantly promotes the practical value of deep learning models and is attracting intense research interests.

\begin{figure}[t]
  \centering  
   \includegraphics[width=\linewidth]{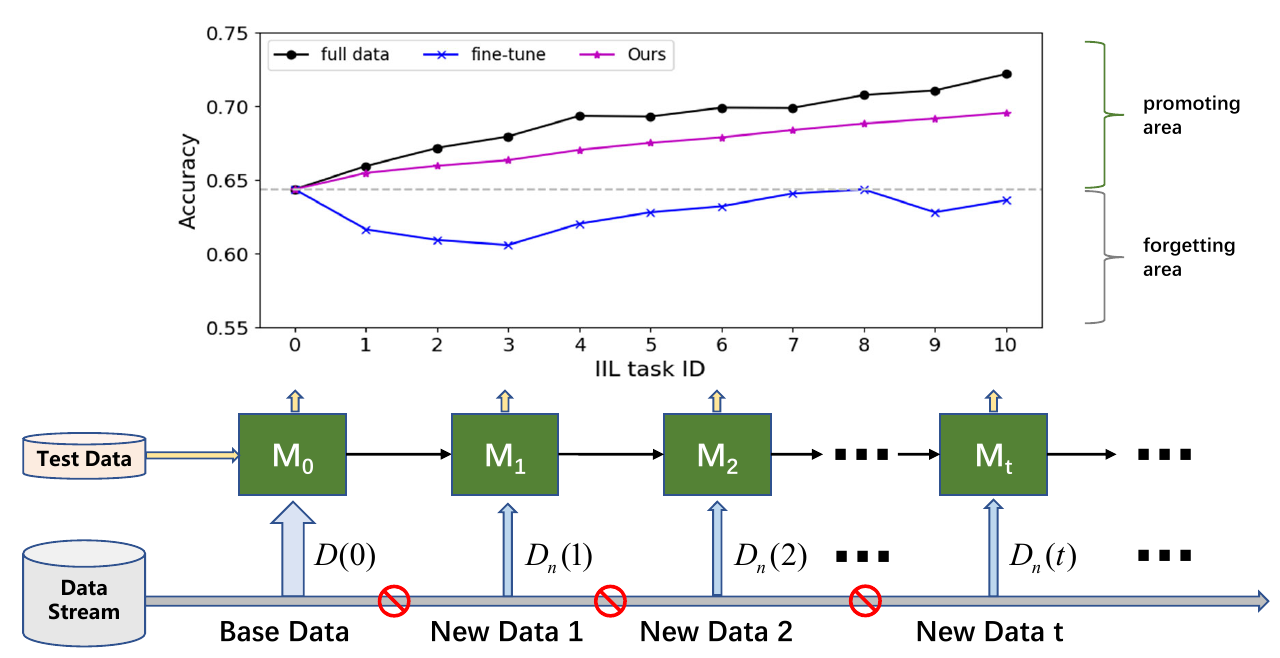}
   \caption{Illustration of the new IIL setting. At the IIL learning phase $t>0$, only the new data $D_n(t)$ that is much smaller than the base data is available. Model should be promoted by only leveraging the new data each time and seeks a performance close to the \textit{full data} model trained on all accumulated data. \textit{Fine-tuning} with early stopping fails to enhance the model in the new IIL setting.}
   \label{fig:problemsetting}
\end{figure}

According to whether the new data comes from seen classes, incremental learning can be divided into three scenarios~\cite{lomonaco2017core50,luo2020appraisal}: instance-incremental learning (IIL)~\cite{de2021continual,lomonaco2017core50} where all new data belongs to the seen classes, class-incremental learning (CIL)~\cite{rebuffi2017icarl, douillard2020podnet, li2017learning, liu2021adaptive} where new data has different class labels, and hybrid-incremental learning~\cite{he2020incremental,wu2022class} where new data consists of new observations from both old and new classes. Compare to CIL, IIL is relatively unexplored because it is less susceptible to the CF. Lomonaco and Maltoni~\cite{lomonaco2017core50} reported that fine-tuning a model with early stopping can well tame the CF problem in IIL. However, this conclusion not always holds when there is no access to the old training data and the new data has a much smaller size than old data, as depicted in Fig.~\ref{fig:problemsetting}. Fine-tuning often results in a shift in the decision boundary rather than expanding it to accommodate new observations. Besides retaining old knowledge, the real deployment concerns more on efficient model promotion in IIL. For instance, in the defect detection of industry products, classes of defect are always limited to known categories. But the morphology of those defects is varying time to time. Failures on those unseen defects should be corrected timely and efficiently to avoid the defective products flowing into the market. Unfortunately, existing research primarily focuses on retaining knowledge on old data rather than enriching the knowledge with new observations.

In this paper, to fast and cost-effective enhance a trained model with new observations of seen classes, we first define a new IIL setting as \textit{retaining the learned knowledge as well as promoting the model's performance on new observations without access to old data}. In simple words, we aim to promote the existing model by only leveraging the new data and attain a performance that is comparable to the model retrained with all accumulated data. The new IIL is challenging due to the concept drift~\cite{he2020incremental} caused by the new observations, such as the color or shape variation compared to the old data. Hence, two issues have to be tackled in the new IIL setting: 1) the notorious catastrophic forgetting because of no access to old data, and 2) broadening the existing decision boundary to new observations. 



To address above issues in the new IIL setting, we propose a novel IIL framework based on the teacher-student structure. The proposed framework consists
of a decision boundary-aware distillation (DBD) process and a knowledge consolidation (KC) process. 
The DBD allows the student model to learn from new observations with awareness of the existing inter-class decision boundaries, which enables the model to determine where to strengthen its knowledge and where to retain it. However, the decision boundary is untraceable when there are insufficient samples located around the boundary because of no access to the old data in IIL. 
To overcome this, we draw inspiration from the practice of dusting the floor with flour to reveal hidden footprints. Similarly, we introduce random Gaussian noise to pollute the input space and manifest the learned decision boundary for distillation. During training the student model with boundary distillation, the updated knowledge is further consolidate back to the teacher model intermittently and repeatedly with the EMA mechanism~\cite{tarvainen2017mean}. Utilizing teacher model as the target model is a pioneering attempt and its feasibility is explained theoretically.

According to the new IIL setting, we reorganize the training set of some existing datasets commonly used in CIL, such as Cifar-100~\cite{krizhevsky2009learning} and ImageNet~\cite{russakovsky2015imagenet}
to establish the benchmarks. Model is evaluated on the test data as well as the non-available base data in each incremental phase.
Our main contributions can be summarized as follows: \textbf{1)} We define a new IIL setting to seek for fast and cost-effective model promotion on new observations and establish the benchmarks; \textbf{2)} We propose a novel decision boundary-aware distillation method to retain the learned knowledge as well as enriching it with new data; \textbf{3)} We creatively consolidate the learned knowledge from student to teacher model to attain better performance and generalizability, and prove the feasibility theoretically; and \textbf{4)} Extensive experiments demonstrate that the proposed method well accumulates knowledge with only new data while most of existing incremental learning methods failed.
\section{Related works}
\label{sec:relatedworks}
This paper devotes to the instance-incremental learning which is an associated topic to the CIL but seldom investigated. In the following, related topics on class-incremental learning, continual domain adaptation, and methods based on knowledge distillation (KD) are introduced.

{\bf Class-incremental learning.} CIL is proposed to learn new classes without suffering from the notorious catastrophic forgetting problem and is the main topic that most of works focused on in this area. Methods of CIL can be categorized into three types: 1) important weights regularization~\cite{aljundi2018memory, kirkpatrick2017overcoming, paik2020overcoming, zenke2017continual}, which constrains the important weights for old tasks and free those unimportant weights for new task. Freezing the weights limits the ability to learn from new data and always lead to a inferior performance on new classes. 2) Rehearsal or pseudo rehearsal method, which stores a small-size of typical exemplars~\cite{castro2018end, douillard2020podnet, kang2022class, rebuffi2017icarl} or relies on a generation network to produce old data~\cite{rios2018closed} for old knowledge retaining. Usually, these methods utilize knowledge distillation and perform over the weight regularization method. \textit{Although the prototypes of old classes are efficacy in preserving knowledge, they are unable to promote the model's performance on hard samples, which is always a problem in real deployment}. 3) Dynamic network architecture based method~\cite{hyder2022incremental, liu2021adaptive, yan2021dynamically, wu2022class}, which adaptively expenses the network each time for new knowledge learning. However, deploying a changing neural model in real scenarios is troublesome, especially when it goes too big. Although most CIL methods have strong ability in learning new classes, few of them can be directly utilized in the new IIL setting in our test. The reason is that performance promotion on old classes is less emphasized in CIL.

{\bf Knowledge distillation-based incremental learning.} Most of existing incremental learning works utilize knowledge distillation (KD) to mitigate catastrophic forgetting. LwF~\cite{li2017learning} is one of the earliest approaches that constrains the prediction of new data through KD. iCarl~\cite{rebuffi2017icarl} and many other methods distill knowledge on preserved exemplars to free the learning capability on new data. Zhai et al.~\cite{zhai2019lifelong} and Zhang et al.~\cite{zhang2020class} exploit distillation with augmented data and unlabeled auxiliary data at negligible cost.
Different from above distillation at label level, Kang et al.~\cite{kang2022class} and Douillard~\cite{douillard2020podnet} proposed to distill knowledge at feature level for CIL. Compared to the aforementioned researches, the proposed decision boundary-aware distillation requires no access to old exemplars and is simple but effective in learning new as well as retaining the old knowledge.

{\bf Comparison with the CDA and ISL.} Rencently, some work of continual domain adptation (CDA)~\cite{hsu2018re,pu2021lifelong,tao2020bi} and incremental subpopulation learning (ISL)~\cite{liang2022balancing} is proposed and has high similarity with the IIL setting. All of the three settings have a fixed label space. The CDA focus on solving the visual domain variations such as illumination and background. 
ISL is a specific case of CDA and pays more attention to the subcategories within a class, such as Poodles and Terriers. Compared to them, IIL is a more general setting where the concept drift is not limited to the domain shift in CDA or subpopulation shifting problem in ISL. More importantly, the new IIL not only aims to retain the performance but also has to promote the generalization with several new observations in the whole data space.

\section{Problem setting}
\label{probsettings}
Illustration of the proposed IIL setting is shown in Fig.~\ref{fig:problemsetting}. As can be seen, data is generated continually and unpredictably in the data stream. Generally in real application, people incline to collect enough data first and train a strong model $M_0$ for deployment. No matter how strong the model is, it inevitably will encounter out-of-distribution data and fail on it. These failed cases and other low-score new observations will be annotated to train the model from time to time. Retraining the model with all cumulate data every time leads to higher and higher cost in time and resource. Therefore, the new IIL aims to enhance the existing model with only the new data each time.

\begin{figure}[t]
  \centering  
   \includegraphics[width=0.85\linewidth]{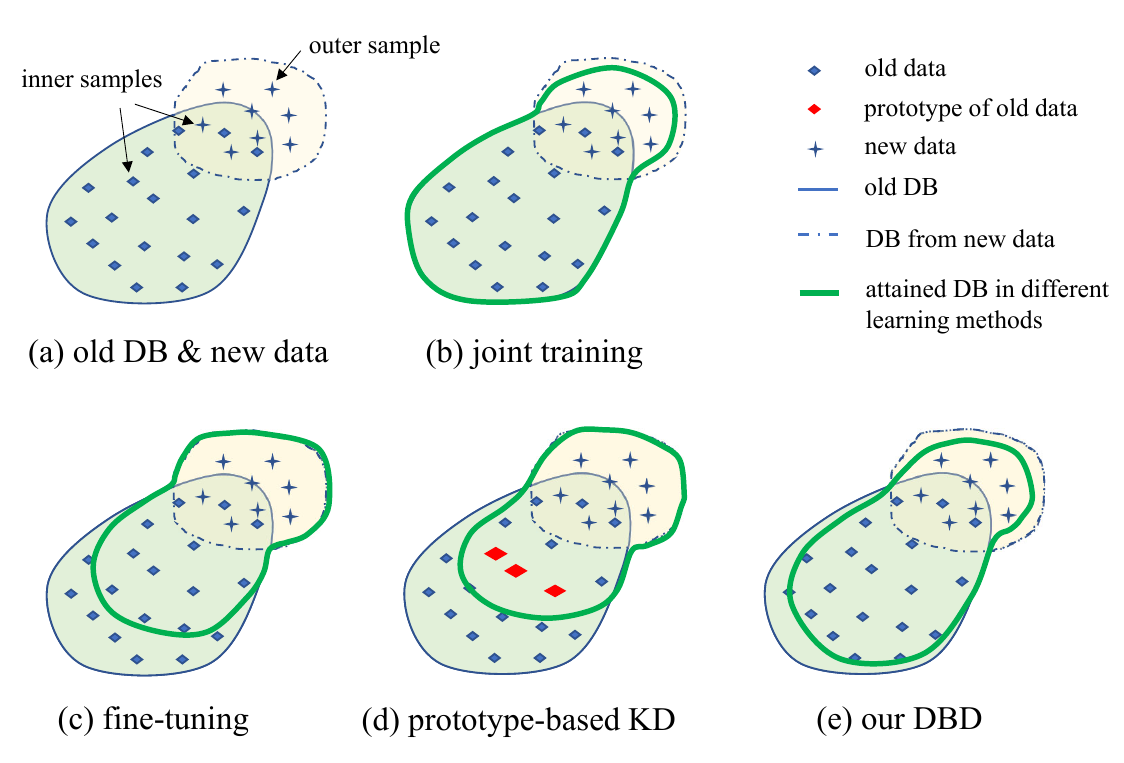}
   \caption{Decision boundaries (DB): (a) DB learned from old data and new data, respectively. With respect to the old DB, new data can be categorized into inner samples and outer samples. (b) ideal DB by jointly training on the old data and new data. (c) fine-tuning the model on the new data with one-hot labels suffers to CF. (d) learning with distillation on prototype exemplars causes overfitting to these exemplars and DB collapsing. (e) the DB achieved using our decision boundary-aware distillation (DBD).}
   \label{fig:DB}
\end{figure}

In the proposed IIL, the classes $C$ are known and fixed. Given a well-trained model $M_{t-1}$  and new observed instances $D_n(t)=\left\{(x_i, y_i): y_i\in C\right\}_{i=1}^N$ at incremental phase $t\geq 1$, the goal is to enhance the model $M_{t-1}$ without any data seen before phase $t$ so that the model can work well both on the new observations and all the data seen before the current phase $t$, \ie better generalization. Therefore, the attained model $M_t$ should enlarge its decision boundary to the misclassified outer samples in the new data as well as retaining the learned boundary on those correctly classified inner samples, as shown in Fig.~\ref{fig:DB}. To facilitate the real application, no structural variation on the model is allowed during learning. That is, $M_t$ and $M_{t-1}$ should have the same structure and number of parameters. The model $M_{t-1}$ can be a strong base model $M_0$ trained on the base dataset $D(0)$ or an incremental model derived from $M_0$ after $t-1$ learning phases. Usually, $D_n(t)$ is much smaller than the size of $D(0)$, i.e., $D_n(t) \ll D(0)$. To evaluate $M_t$, a fixed test dataset $D_{test}$ is used and $M_t$ should have better performance on it after learning on the new data each time.

\section{Methodology}
As shown in Fig.~\ref{fig:DB} (a), the occurrence of concept drift in new observations leads to the emergence of outer samples that the existing model fails on. The new IIL has to broaden the decision boundary to these outer samples as well as avoiding the catastrophic forgetting (CF) on the old boundary. Conventional knowledge distillation-based methods rely on some preserved exemplars~\cite{rebuffi2017icarl} or auxiliary data~\cite{zhai2019lifelong, zhang2020class} to resist CF. However, in the proposed IIL setting, we have no access to any old data other than new observations. Distillation based on these new observations conflicts with learning new knowledge if no new parameters are added to the model. To strike a balance between learning and not forgetting, we propose a decision boundary-aware distillation method that requires no old data. 
During learning, the new knowledge learned by the student is intermittently consolidated back to the teacher model, which brings better generalization and is a pioneer attempt in this area. 

\subsection{Decision boundary-aware distillation}
\label{sec:dbd}
Decision boundary (DB) which reflects the inter-class relationship and intra-class distribution is one of the most valuable knowledge stored in a well-trained model. It can be defined by distinguishing between inner samples (correctly classified) and outer samples (misclassified), as illustrated in \cref{fig:DB} (a). In new IIL, promoting the model's performance on new data without forgetting equals to extend the existing DB to enclose those new outer samples while retain the DB in other locations. However, to learn from new data, existing methods take the annotated one-hot labels as the optimal learning target for granted. We argue that one-hot labels ignore the relationship between target class and other classes. Naively learning with one-hot labels tends to push outer samples towards the DB center, which can potentially interfere with the learning of other classes, especially when the data is insufficient to rectify such interference in IIL. 

\begin{figure}[t]
  \centering  
   \includegraphics[width=\linewidth]{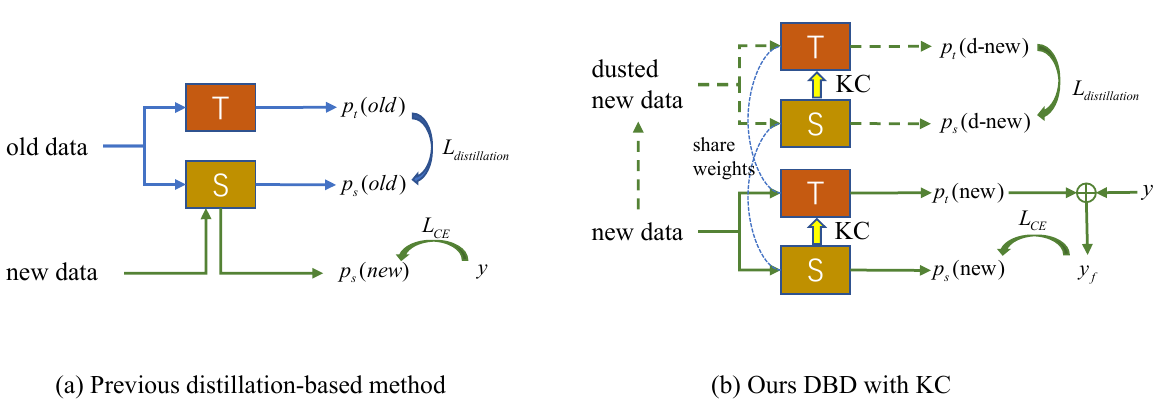}
   \caption{Comparison between (a) previous distillation-based method which inferences with student model (S) and (b) the proposed decision boundary-aware distillation (DBD) with knowledge consolidation (KC). We use teacher model (T) for inference.}
   \label{fig:method}
\end{figure}

To address the inter-class interference, we propose to learn the new data by fusing annotated one-hot labels with predictions of existing model, as show in \cref{eq:smooth_l}.  
For new outer samples, learning with fused labels retains the knowledge on none-target classes and extends the decision boundary more moderately to enclose them. Knowledge of none-target classes is crucial for retain learned knowledge, which also reported by Mittal \etal~\cite{mittal2021essentials} in using super-class labels. For new inner samples, keeping the DB around it is a safer choice, \ie using existing predicted scores as the learning target. However, we propose to push the DB away from the new peripheral inner samples by sharpening the teacher's prediction score with the one-hot label, which in essence enlarges the inter-class distance while retaining the DB. Hence, although with different motivation, training on new outer samples and new inner samples are unified through the fused label.
\begin{equation}\label{eq:smooth_l}
    y_{f}^k(x) = \frac{(y^k + p_t^k)/\tau}{\sum_j (y^j + p_t^j)/\tau}
\end{equation}


\noindent where $y_{f}(x)$ denotes the fused label of sample $x\in X$, $k,j \in C$, $y$ is the annotated one-hot label, $p_t$ is the prediction of teacher model, $\tau$ is the temperature.

Alternating the learning target with fused labels unifies the knowledge learning and retaining on new samples in a simple manner. Although this manner helps retain some learned knowledge, preservation of DB in other directions that only supported by the old data is not addressed. 

The learned DB becomes indistinct when the old data is no longer available after training. In other words, the DB can only be accurately known when a sufficient number of samples are observed in the peripheral area of the DB, which pose a challenge when training with small-size incremental data. As shown in \cref{fig:DB} (c) and (d), learning with a small-size new data inevitably lead to the old DB collapsing, \ie forgetting. To address this challenge, we draw inspiration from a common approach in daily life where people use white powder to make hard-to-see trajectories or footprints visible. In a similar vein, we propose a simple yet effective method to further manifest the learned DB by dusting the input space. Specifically, we add strong random Gaussian noise as perturbations to the input, aiming to relocate some samples to the peripheral area of the DB. Dusting the input space (DIS) is inherently different from the data augmentation which aims for better feature learning and robustness to intra-class variations. DIS target to find and distill the learned boundary in new tasks by pushing samples into the peripheral area where predictions on dusted samples can be different from their original annotated labels. DIS can be formulated by \cref{eq:input}, where $X \in R^{h\times w \times 3}$ denotes the set of input images, $G \in R^{h\times w \times 3}$ is the Gaussian noise with a mean $\mu$ and a standard deviation $\delta$. $norm$ denotes the normalizing operation. 
\begin{equation}\label{eq:input}
    X' = norm(X) + G(\mu, \delta) 
\end{equation}

The whole DBD is depicted in \cref{fig:method} and the final training loss $\mathcal{L}_{cls}$ is shown in \cref{eq:train}, where $\lambda$ is a weight factor to balance the learning on new samples and the distillation on dusted samples. \\
\begin{multline}\label{eq:train}
    \mathcal{L}_{cls} = -\frac{1}{N}\left(\underbrace{\sum_{x\sim X} y_{f}(x)\log p_s(x)}_{\rm learning \& distillation}+ \right.\\
            \left. \lambda\underbrace{\sum_{x'\sim X'} p_t(x')\log p_s(x')}_{\rm distillation} \right) 
\end{multline}
           

Decision boundary-aware distillation enables the student network to learning new knowledge with awareness of the existing knowledge. 

\subsection{Knowledge consolidation}
\label{sec:rkc}
Different from existing IIL methods that only focus on the student model, we propose to consolidate knowledge from student to teacher for better balance between learning and forgetting. The consolidation is not implemented through learning but through model exponential moving average (EMA). Model EMA was initially introduced by Tarvainen \etal~\cite{tarvainen2017mean} to enhance the generalizability of models. In the vanilla model EMA, the model is trained from scratch, and EMA is applied after every iteration. The underlying mechanism of model EMA is not thoroughly explained before. In this work, we leverage model EMA for knowledge consolidation (KC) in the context of IIL task and explain the mechanism theoretically. According to our theoretical analysis, we propose a new KC-EMA for knowledge consolidation. 

Mathematically, the model EMA can be formulated as \cref{eq:EMA}, where $\theta_t^n$ denotes the weights of teacher model at step $n$ and $\theta_s^n$ is the weights of student model. $n$ is the EMA steps. $\alpha \in [0,1]$ is the EMA momentum.
\begin{equation}\label{eq:EMA}
    \theta_t^n = \alpha \theta_t^{n-1} + (1-\alpha) \theta_s^n
\end{equation}

Based on \cref{eq:EMA}, we can achieve a relationship between $\theta_t^0$ and $\theta_s^n$ as:
\begin{multline}\label{eq:tweight}
    \theta_t^n = \alpha^n \theta_t^0 + \alpha^{n-1}(1-\alpha)\theta_s^1 + \cdots \\ + \alpha(1-\alpha)\theta_s^{n-1} + (1-\alpha)\alpha_s^n
\end{multline}

Noted that $\theta_t^0=\theta_s^0$ in IIL. As the base model is well trained and converges, we have $\frac{d\mathcal{L}_{old}}{d\theta_t^0}\approx0$, where $\mathcal{L}_{old}$ denotes the training loss on the old data. After thoroughly training the student model on new incremental data, the student model converges on the new observations and we have $\frac{d\mathcal{L}_{new}}{d\theta_s^n}\approx0$. Based on \cref{eq:tweight}, the derivative of old task and new task on the attained teacher weights is 
\begin{equation}\label{eq:td}
    \frac{d(\mathcal{L}_{old}+\mathcal{L}_{new})}{d\theta_t^n} = \frac{1}{\alpha^n} \cdot \frac{d\mathcal{L}_{old}}{d\theta_t^0} + \frac{1}{1-\alpha}\cdot \frac{d\mathcal{L}_{new}}{d\theta_s^n}
\end{equation}

When $n$ is not sufficiently large, as $\frac{d\mathcal{L}_{old}}{d\theta_t^0}\approx0$ and $\frac{d\mathcal{L}_{new}}{d\theta_s^n}\approx0$, we can get that $\frac{d(\mathcal{L}_{old}+\mathcal{L}_{new})}{d\theta_t^n} \approx 0$.

Hence, the teacher model can achieve a minima training loss on both the old task and the new task, which indicates improved generalization on both the old data and new observations. This has been verified by our experiments in \cref{sec:exp}. However, since $\alpha < 1$, it is noteworthy that the gradient of the teacher model, whether on the old task or the new task, is larger than the initial gradient on the old task or the final gradient of the student model on the new task. That is, the obtained teacher model sacrifices some unilateral performance on either the old data or the new data in order to achieve better generalization on both. From this perspective, the mechanism of vanilla EMA could also be partially explained. In vanilla EMA, where the model starts from scratch and only the new task is considered, we only need to focus on the second term in Equation \ref{eq:td}. Since the teacher model has larger gradient on the training data than the student model, it is less possible to overfit to the training data. As a result, the teacher model has better generalization as Tarvainen \etal~\cite{tarvainen2017mean} observed.

Moreover, the term $\frac{1}{\alpha^n}$ is inversely proportional to $\frac{1}{1-\alpha}$ when $n$ is fixed, which implies that decreasing the teacher's gradient on the new data by reducing $\alpha$ will cause an increase in the gradient on the old data. In simple, making the teacher model closer to the student model damages the performance on the old task. This observation aligns with general expectations. However, it also indicates that  there is a trade-off between the performance on new observations and the performance on the old data in knowledge consolidation. This trade-off results in an inevitable gap between the teacher model and the full-data model. The fact that the student model is already trained to perform well on both the old data and new data with DBD alleviates such a contradiction and facilitates consolidating the knowledge to teacher for better performance.  

To prevent large EMA steps $n$ from causing serious CF, KC-EMA is performed every 5 epochs after the initial 10 epochs of training. The updating momentum $\alpha$ is also designed to adaptively increase its weight on the student model along with epoch $e$, as shown in~\cref{eq:momentum}. $e_w$ is a warm up value. This approach guarantees that KC occurs when student model is thoroughly trained. Updating the teacher weights too frequent harms retaining the previously learned knowledge, while too infrequent updating hinders the student model from learning new knowledge. In essence, the student model plays a role of quick learner, while the teacher is a slow learner to balance between tasks.

\begin{equation}\label{eq:momentum}
    \alpha = min(\alpha_0, 1-\frac{e}{e+e_w})
\end{equation}
\section{Experimental results}
\label{sec:exp}
We reorganize the training set of some existing datasets that are commonly used in the class-incremental learning to establish the benchmarks. Implementation details of our experiments can be found in the supplementary material.

\begin{table*}[ht]
    \centering
    \renewcommand{\arraystretch}{1.1}
    \resizebox{0.85\linewidth}{!}{
    \begin{tabular}{ccc|cccc|cccc}
    \hline
       \multicolumn{3}{c|}{\multirow{2}{*}{Method}} & \multicolumn{4}{c|}{Cifar-100} & \multicolumn{4}{c}{ImageNet-100} \\
    \cline{4-11}
       & & & $P_0$ (\%) & $P_{10}$ (\%) & {$PP$ (\%)} & {$\mathcal{F}$ (\%)} & $P_0$ (\%) & $P_{10}$ (\%) & {$PP$(\%) } & {$\mathcal{F}$ (\%)} \\
    \cline{4-11}
       \hline
       \multirow{2}{*}{baseline}  & Full data & & 64.34 & 71.44$_{\pm0.42}$ & +7.1$_{\pm0.42}$ & -0.13$_{\pm0.05}$ & 71.54 & 78.24$_{\pm0.23}$ & +6.7$_{\pm0.23}$ &+1.12$_{\pm0.09}$ \\
       & Fine-tuning & &64.34 &64.53$_{\pm0.62}$ & +0.19$_{\pm0.62}$ &-9.03$_{\pm0.68}$ &71.54 &68.74$_{\pm0.17}$ & -2.8$_{\pm0.17}$ &-15.41$_{\pm0.20}$ \\
       \hline
       \multirow{4}{*}{rehearsal-based}  & iCarl~\cite{rebuffi2017icarl} & CVPR'17 & 66.20 & 61.61$_{\pm0.48}$ & -4.59$_{\pm0.48}$  & -33.06$_{\pm0.47}$ & 72.2 & 72.3$_{\pm0.22}$ & +0.1$_{\pm0.22}$ &-11.58$_{\pm0.11}$ \\
       & PoDNet~\cite{douillard2020podnet} & ECCV'20 & 65.1 & 61.08$_{\pm0.16}$ & -4.02$_{\pm0.16}$ & -20.68$_{\pm0.14}$ & 76.00 & 72.36$_{\pm0.34}$ & -3.64$_{\pm0.34}$ &-11.84$_{\pm0.13}$  \\       
       & Der~\cite{yan2021dynamically} & CVPR'21 & 64.80 & 62.10$_{\pm0.26}$ & -2.70$_{\pm0.26}$ & -21.40$_{\pm0.12}$ & 74.30 & 71.20$_{\pm0.28}$ & -3.10$_{\pm0.28}$ & -17.44$_{\pm0.10}$ \\
       & OnPro~\cite{wei2023online} & ICCV'23 & 65.12 & 61.44$_{\pm0.19}$ & -3.69$_{\pm0.19}$ & -7.04$_{\pm0.12}$ & 69.04 & 66.90$_{\pm0.24}$ & -2.14$_{\pm0.24}$ & -3.37$_{\pm0.09}$ \\
       \hline
       \multirow{4}{*}{exemplar-free} & LwF~\cite{li2017learning} & TPAMI'17 & 64.34 & 67.17$_{\pm0.26}$ & +2.83$_{\pm0.26}$ & -7.98$_{\pm0.95}$ & 71.54 & 69.75$_{\pm0.36}$ & -1.79$_{\pm0.36}$ & -16.12$_{\pm0.10}$ \\       
       & online learning~\cite{he2020incremental} & CVPR'20 & 64.34 & 64.70$_{\pm0.18}$ & +0.36$_{\pm0.18}$ & \bf{-1.48$_{\pm0.08}$} & 71.54 & 68.55$_{\pm0.31}$ & -2.99$_{\pm0.31}$ & -12.73$_{\pm0.32}$ \\
       & ISL~\cite{liang2022balancing} & ECCV'22  & 65.48 & 66.23$_{\pm0.13}$ & +0.75$_{\pm0.13}$ & -2.76$_{\pm0.13}$ & 71.54 & 70.99$_{\pm0.33}$ & -0.55$_{\pm0.33}$ & \bf{-1.92$_{\pm0.12}$} \\
       & \bf{Ours} & & 64.34 & \bf{69.27$_{\pm0.21}$} & \bf{+4.93$_{\pm0.21}$} & -1.86$_{\pm0.08}$ & 71.54 & \bf{73.87$_{\pm0.13}$} & \bf{+2.33$_{\pm0.13}$} & -5.79$_{\pm0.06}$ \\
    \hline
    \end{tabular}}
    \caption{Instance-incremental learning on Cifar-100 and ImageNet.The ${PP}$ reflects the accuracy changing on test data $D_{test}$ over 10 IIL tasks. $\mathcal{F}$ is the forgetting rate on base training data $D(0)$ after last IIL task. Results are average score and their 95\% confidence interval of 5 runs with different incremental data orders. Following previous works, resnet-18 is used as the backbone network for all experiments.}
    \label{tab:sota_com}
\end{table*}

\begin{figure*}[t]
    \centering
    \includegraphics[width=\linewidth]{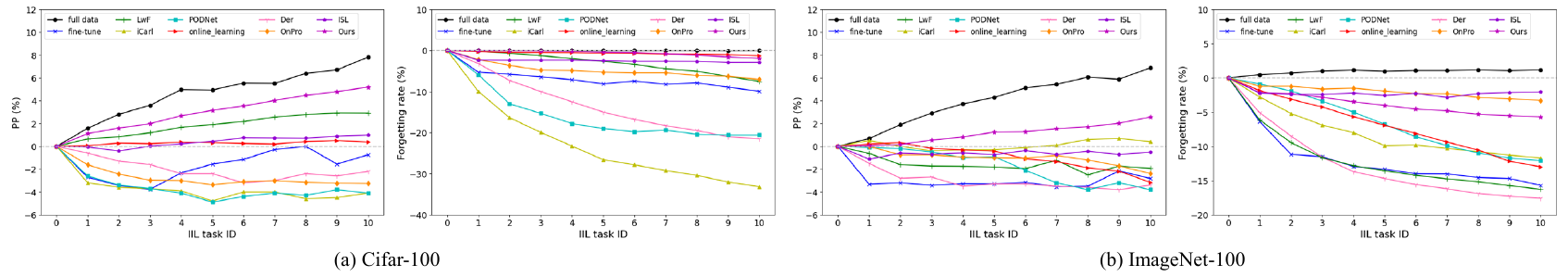}
    \caption{Detailed performance promotion ($PP$) and forgetting rate ($\mathcal{F}$) at each IIL phase. Best to view in color with scaling.}
    \label{fig:PP}
\end{figure*}
\subsection{Experiment Setup}
\subsubsection{Datasets}
\textbf{Cifar-100}~\cite{krizhevsky2009learning} is the most commonly used dataset in incremental learning. It contains 60K $32\times32$ images of 100 classes. Each class has 500 images for training and 100 images for testing. In CIL, the training set is further split to many small subsets in terms of class. In the new IIL, we randomly choose 25K images as the base dataset $D(0)$ to train a base model $M_0$. The remaining 25K images are randomly divided into 10 incremental sets. Thus, $D_n(t)=\frac{1}{10}D(0)$. Each incremental set contains samples from all classes but might has different number for each class. 
The test set of 10K images is used for evaluation after each learning phase.

\textbf{ImageNet}~\cite{russakovsky2015imagenet} is another dataset that commonly used. The ImageNet-1000 which consists of 1.2 million training images and 150K testing images from 1000 classes. Following Douillard \etal~\cite{douillard2020podnet, he2020incremental}, we randomly select 100 classes (ImageNet-100) and split it into 1 base set with half of the training images and 10 incremental sets with another half of images as we do on Cifar-100.  

\textbf{Entity-30} included in BREEDS datasets~\cite{santurkar2021breeds} simulates the real-world sub-population shifting. For example, the base model learns the concept of dog with photos of ``Poodles", but on incremental data it has to extend the ``dog" concept to ``Terriers" or ``Dalmatians". Entity-30 has 240 subclasses with a large data size. 
As the sub-population shifting is a specific case of the instance-level concept drift, we evaluate the proposed method on Entity-30 following the setting of ISL~\cite{liang2022balancing}.
\subsubsection{Evaluation metrics}
After each incremental learning phase, the model is evaluated on the test set $D_{test}$ to assess the performance improvement achieved by each incremental model as well as on the no longer available base data $D_0$ to measure the extent of forgetting. 
As different methods may have different base performance on $M_0$, performance promotion (${PP}$) over $T$ IIL tasks on $D_{test}$ is used for comparison between methods, as shown in \cref{eq:pp}. $P_t(D)$ denotes the performance of model $M_t$ on data $D$. Forgetting rate $\mathcal{F}$ after the last learning phase is reported as the accuracy degradation of $M_T$ on $D_0$ when contrasted to $M_0$ which fully trained on $D_0$, as shown in \cref{eq:forgetting}. In \cref{tab:sota_com}, performance of $M_t$ on $D_{test}$ is denoted as $P_t$ for short.
\begin{align}   
    &{PP} = \sum_{1\leq t \leq T} [P_{t}(D_{test}) - P_{t-1}(D_{test})] \label{eq:pp} \\
    &\mathcal{F} = P_T(D_0) - P_0(D_0)  \label{eq:forgetting}  
\end{align}


\subsubsection{Evaluated baselines}
As few existing method is proposed for the IIL setting, we reproduce several classic and SOTA CIL methods by referring to their original code or paper with the minimum revision, including iCarl~\cite{rebuffi2017icarl} and LwF~\cite{li2017learning} which utilize label-level distillation,  PODNet~\cite{douillard2020podnet} which implements distillation at the feature level, Der~\cite{yan2021dynamically} which expends the network dynamically and attains the best CIL results, OnPro~\cite{wei2023online} which uses online prototypes to enhance the existing boundaries, and online learning~\cite{he2020incremental} which can be applied to the hybrid-incremental learning. ISL~\cite{liang2022balancing} proposed for incremental sub-population learning is the only method that can be directly implemented in the new IIL setting. 
As most CIL methods require old exemplars, to compare with them, we additionally set a memory of 20 exemplars per class for these methods. We aim to provide a fair and comprehensive comparison in the new IIL scenario. Details of reproducing these methods can be found in our supp. material.

To illustrate the effect of our methods, we also report three baseline results: full data, fine-tuning, and vanilla distillation. ``Full data" represents the case training with all accumulated data at each phase. If without specific illustration, ``fine-tune" means fine-tuning for 10 epochs only with new data, \ie, fine-tuning with early stopping~\cite{lomonaco2017core50}. In ``vanilla distillation" as shown in \cref{fig:method} (a), we randomly select 10\% new data $\mathcal{E} \subset D_n(t)$ for knowledge distillation and re-sample the data in $\mathcal{E}$ and $D_n(t)-\mathcal{E}$ to form a balance mini-batch as previous distillation-based methods do.
\subsection{Comparison with SOTA methods}
\cref{tab:sota_com} shows the test performance of different methods on the Cifar-100 and ImageNet-100. The proposed method achieves the best performance promotion after ten consecutive IIL tasks by a large margin with a low forgetting rate. Although ISL~\cite{liang2022balancing} which is proposed for a similar setting of learning from new sub-categories has a low forgetting rate, it fails on the new requirement of model enhancement. Attain a better performance on the test data is more important than forgetting on a certain data.

In the new IIL setting, all rehearsal-based methods including iCarl~\cite{rebuffi2017icarl}, PODNet~\cite{douillard2020podnet}, Der~\cite{yan2021dynamically} and OnPro~\cite{wei2023online}, not perform well. Old exemplars can cause memory overfitting and model bias~\cite{zhangsimple}. Thus, limited old exemplars not always have a positive influence to the stability and plasticity~\cite{sunexploring}, especially in the IIL task. 
Forgetting rate of rehearsal-based methods is high compared to other methods, which also explains their performance degradation on the test data. Detailed performance at each learning phase is shown in \cref{fig:PP}. Compared to other methods that struggle in resisting forgetting, our method is the only one that stably promotes the existing model on both of the two datasets.

Following ISL~\cite{liang2022balancing}, we further apply our method on the incremental sub-population learning as shown in \cref{tab:isl}. Sub-population incremental learning is a special case of the IIL where new knowledge comes from the new subclasses. Compared to the SOTA ISL~\cite{liang2022balancing}, our method is notably superior in learning new subclasses over long incremental steps with a comparable small forgetting rate. Noteworthy, ISL~\cite{liang2022balancing} use Continual Hyperparameter Framework (CHF)~\cite{de2021continual} searching the best learning rate (such as low to 0.005 in 15-step task) for each setting. While our method learns utilizing ISL pretrained base model with a fixed learning rate (0.05). Low learning rate in ISL reduces the forgetting but hinders the new knowledge learning. The proposed method well balances learning new from unseen subclasses and resisting forgetting on seen classes.

\begin{table*}
    \centering
    \begin{tabular}{c|ccc|ccc|ccc}
         \hline
         \multirow{2}{*}{Method} &  \multicolumn{3}{c|}{4 Steps}&  \multicolumn{3}{c|}{8 Steps}&  \multicolumn{3}{c}{15 Steps}\\
         \cline{2-10}
          &  Unseen($\uparrow$)&  All($\uparrow$)& $\mathcal{F}_4 $($\downarrow$)&  Unseen($\uparrow$)&  All($\uparrow$)& $\mathcal{F}_8$($\downarrow$) & Unseen($\uparrow$)&  All($\uparrow$)& $\mathcal{F}_{15}$($\downarrow$)\\  
         \hline
         Full data&  88.03&  87.63&  -&  88.03&  87.63&  -&  88.03&  87.63& -\\
         \hline
         Fine-tuning &  53.72&  48.08&  47.75&  26.45&  23.08&  73.86&  14.68&  13.77& 84.49\\
         EWC~\cite{kirkpatrick2017overcoming}&  56.17&  54.10&  40.69&  30.50&  29.00&  66.94&  22.20&  23.68& 74.03\\
         LwF~\cite{li2017learning}&  62.67&  58.85&  32.32&  34.52&  29.69&  64.38&  32.62&  31.17& 62.51\\
         iCarl~\cite{rebuffi2017icarl}&  68.28&  64.43&  28.20&  46.93&  43.69&  50.88&  34.53&  33.79& 62.36\\
         MUC~\cite{liu2020more}&  62.98&  59.59&  29.45&  36.17&  31.83&  61.49&  34.15&  32.54& 60.65\\
         PASS~\cite{zhu2021prototype}&  64.50&  69.37&  21.79&  48.85&  54.99&  40.50&  32.13&  39.75& 58.27\\
         ISL~\cite{liang2022balancing}& 64.73& 72.88& 4.16& 58.63& 72.14& 2.30 & 56.87& \bf{71.69}&\bf{3.48}\\
         \hline
        \bf{Ours}& \bf{65.17} & \bf{75.68} & \bf{0.004}& \bf{65.88}& \bf{74.89}& \bf{2.23}& \bf{58.85}& 70.25&8.73\\
        \hline
    \end{tabular}
    \caption{Results of incremental sub-population learning on Entity-30 benchmark. \textit{Unseen, All} and $\mathcal{F}_i$ denote the average test accuracy on unseen subclasses in incremental data, on all seen (base data) and unseen subclasses, and the average forgetting rate over all test data. More details of the metrics can be found in ISL~\cite{liang2022balancing}.}
    \label{tab:isl}
\end{table*}
\subsection{Ablation study}
All ablation studies are implemented on Cifar-100 dataset.  

\begin{figure}[t]
    \centering
    \includegraphics[width=0.7\linewidth]{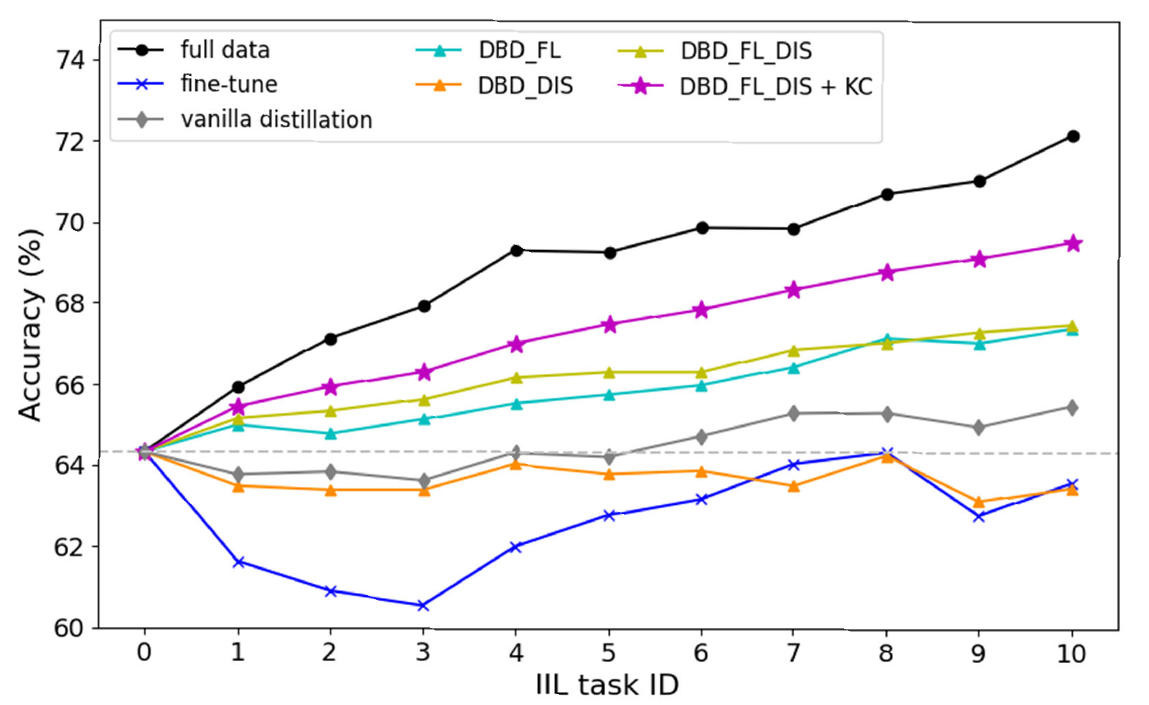}
    \caption{Effect of three components in DBD: fused label (FL), dusted input space (DIS), and knowledge consolidation (KC). It can be seen that all components contributes to the continual knowledge accumulation with new data.}
    \label{fig:components}
\end{figure}

\textbf{Effect of each component.} The proposed method mainly consists of three components: DBD with fused label (FL), DBD with dusting the input space (DIS), and knowledge consolidation (KC). The ablation study on these three components is shown in~\cref{fig:components}. It can be seen that DBD with all components has the largest performance promotion in all phases. Although DBD with only DIS fails to enhance the model (which can be understood), it still shows great potential in resisting CF contrasted to fine-tuning with early stopping. The bad performance of fine-tuning also verifies our analysis that learning with one-hot label causes the decision boundary shifting to other than broadening to the new data. Different from previous distillation base on one-hot label, fused label well balances the need for retaining old knowledge and learning from new observation. Combining the boundary-aware distillation with knowledge consolidation, the model can better tame the knowledge learning and retaining problem with only new data. Consolidating the knowledge to teacher model during learning not only releases the student model in learning new knowledge, but also an effective way to avoid overfitting to new data. 

\begin{figure}[t]
    \centering
    \includegraphics[width=\linewidth]{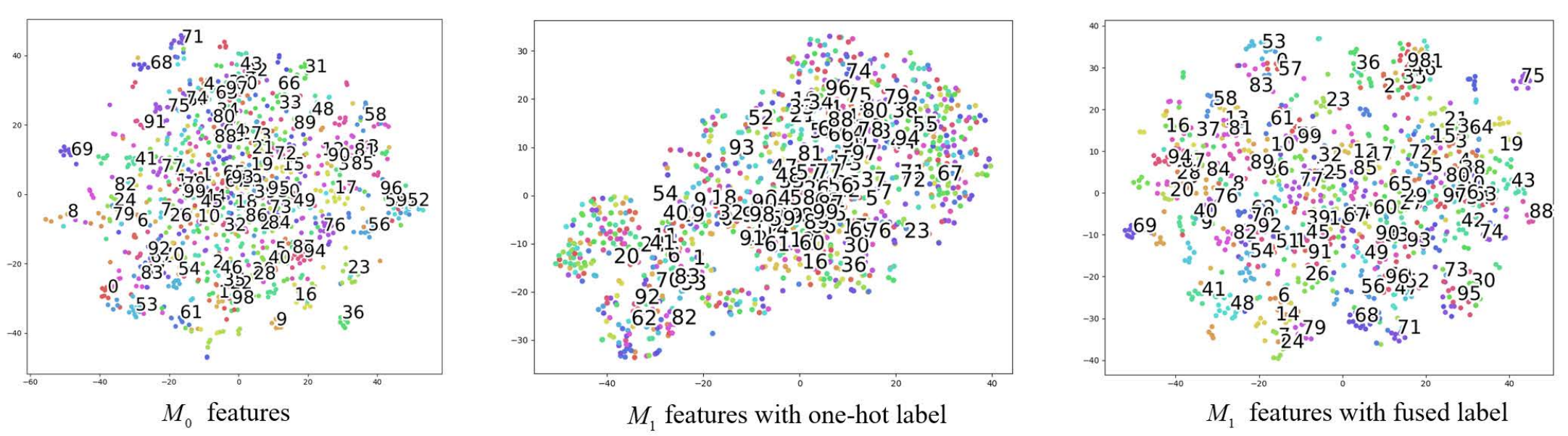}
    \caption{t-SNE visualization of features after first IIL phase, where the numbers denotes classes. 
    }
    \label{fig:features}
\end{figure}

\textbf{Impact of fused labels.} In \cref{sec:dbd}, the different learning demands for new outer samples and new inner samples in DBD are analyzed. 
We propose to use fused label for unifying knowledge retaining and learning on new data. Features learned in the first incremental phase with fused label and one-hot label are contrasted in \cref{fig:features}. Learning with fused label not only retains most of the feature's distribution but also has better separability (more dispersed) compared to the base model $M_0$. While learning with one-hot label changes the feature distribution into a elongated shape. 

\begin{table}[]
    \centering
    \resizebox{\linewidth}{!}{
    \begin{tabular}{c|c|c|c|c|c}
    \hline
           &one-hot label & teacher's score & fused label & ${PP}$ (\%) & $\mathcal{F}$ (\%)\\
        \hline
         \textcolor{blue}{inner samples} & \textcolor{blue}{\checkmark} & & & \multirow{2}{*}{-0.88} & \multirow{2}{*}{-17.57}\\
         \textcolor{red}{outer samples} & \textcolor{red}{\checkmark} & & & \\
        \hline
        \textcolor{blue}{inner samples} &  & \textcolor{blue}{\checkmark} & & \multirow{2}{*}{+0.81} & \multirow{2}{*}{-9.08}\\
         \textcolor{red}{outer samples} & \textcolor{red}{\checkmark} & & & \\
         \hline
         \textcolor{blue}{inner samples} &  & & \textcolor{blue}{\checkmark} & \multirow{2}{*}{+1.69} & \multirow{2}{*}{-10.83}\\
         \textcolor{red}{outer samples} & \textcolor{red}{\checkmark} & & & \\
         \hline\hline
         \textcolor{blue}{inner samples} & \textcolor{blue}{\checkmark} & & & \multirow{2}{*}{+3.77} & \multirow{2}{*}{-8.81}\\
         \textcolor{red}{outer samples} & & &\textcolor{red}{\checkmark} & \\
         \hline
         \textcolor{blue}{inner samples} &  & \textcolor{blue}{\checkmark} & & \multirow{2}{*}{+3.48} & \multirow{2}{*}{-4.53}\\
         \textcolor{red}{outer samples} &  & & \textcolor{red}{\checkmark} & \\
         \hline
         \textcolor{blue}{inner samples} &  & & \textcolor{blue}{\checkmark} & \multirow{2}{*}{\bf{+4.09}} & \multirow{2}{*}{\bf{-3.31}}\\
         \textcolor{red}{outer samples} &  & & \textcolor{red}{\checkmark} & \\
         \hline
    \end{tabular}}
    \caption{Performance of the student network in IIL tasks by assigning different labels to new samples as learning target. }
    \label{tab:diff_label}
\end{table}

\cref{tab:diff_label} shows the performance of student model when applying different learning target to the new inner samples and new outer samples, which reveals the influence of different labels in boundary distillation. As can be seen, utilizing one-hot label for learning new samples degrades the model with a large forgetting rate. Training the inner samples with teacher score and outer samples with one-hot labels is similar with existing rehearsal-based methods which distills knowledge using the teacher's prediction on exemplars and learns from new data with annotated labels. Such a kind of manner reduces forgetting rate but benefits less in learning new. When applying fused labels to outer samples, the student model can well aware the existing DB and has much better performance. Notably, using the fused label for all new samples achieves the best performance. FL benefits retaining the old knowledge as well as enlarging the DB. 

\textbf{Impact of DIS.} To locate and distill the learned decision boundary, we dust the input space with strong Gaussian noises as perturbations. The Gaussian noises are different from the one used in data augmentation because of its high deviation. In data augmentation, the image after strong augmentation should not change its label. However, there is no such limit in DIS. We hope to relocate inputs to the peripheral area among classes other than keeping them in the same category as they are. In our experiments, the Gaussian noises in pixel intensity obey $N(0, 10)$. 
Sensitivity to the noise intensity and related DIS loss are shown in \cref{tab:sensitivity}. It can be seen that when noise intensity is small as used in data augmentation, the promotion is little considering the result of base model is 64.34\%. Best result is attained when the deviation of noise $\delta=10$. When the noise intensity is too large, it might push all input images as outer samples and do no help in locating the decision boundary. Moreover, it will significantly alter the parameters of batch normalization layers in the network, which deteriorates the model. Visualization of the dusted input image can be found in our supplementary material. Different to noise intensity, the model is less sensitive to the DIS loss factor $\lambda$.

\begin{table}[t]
    \centering
    \resizebox{\linewidth}{!}{
    \begin{tabular}{l|c|c|c|c|c|c}
    \hline
       \textbf{Noise intensity} $\delta$ & 0.1 & 1 & 5 & 10 & 20 & 50  \\
       \hline
       Acc. (\%) of $M_1$  & 64.89 &65.01 &64.98 &\bf{65.30} &65.03 &62.34\\
       \hline
       \textbf{Loss factor} $\lambda$ & 0.1 & 0.5 & 1 & 2 & 5 & 10 \\
       \hline
       Acc. (\%) of $M_1$   &\bf{65.43} &65.14 &65.30 &65.09 &65.24 &65.01 \\
    \hline
    \end{tabular}}
    \caption{Analysis of sensitivity to the noise intensity and DIS loss on Cifar-100. The student's performance is reported as it is directly affected by DIS.}
    \label{tab:sensitivity}
\end{table}

\begin{figure}[t]
    \centering
    \includegraphics[width=\linewidth]{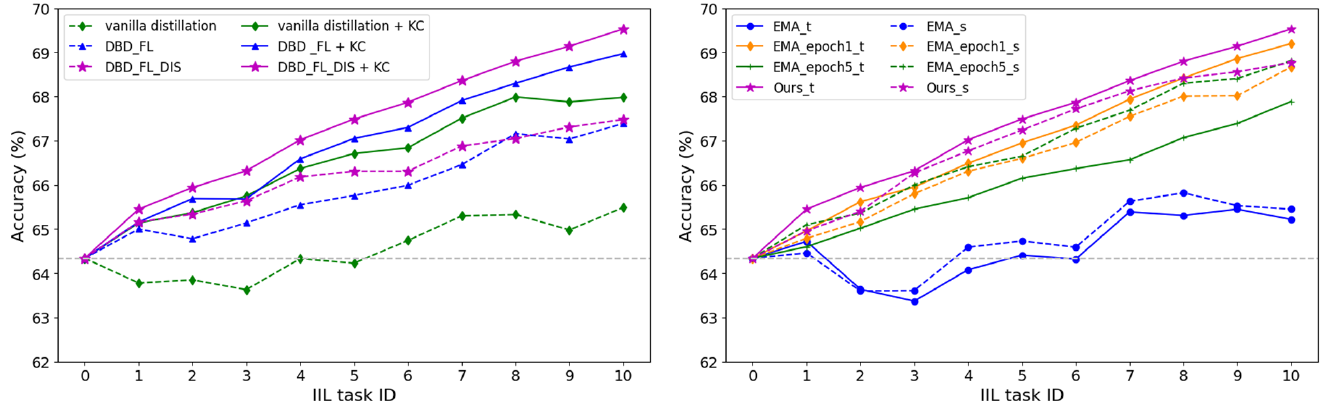}
    \caption{Knowledge consolidation (KC). Left: the solid lines draw the results with KC and the dashed lines are without KC. It shows that the KC significantly promotes the performance. Right: Influence of different model EMA strategies on the teacher (t, solid line) and student (s, dashed line) model. }
    \label{fig:kc}
\end{figure}

\textbf{Impact of knowledge consolidation.} It has been proved theoretically that consolidating the knowledge to teacher model is capable to achieve a model with better generalization on both of the old task and the new task. \cref{fig:kc} left shows the model performance with and without the KC. No matter applied on the vanilla distillation method or the proposed DBD, the knowledge consolidation demonstrates great potential in accumulating knowledge from the student model and promoting the model's performance. However, not all model EMA strategy can works. As shown in \cref{fig:kc} right, traditional EMA that implements after every iteration fails to accumulate knowledge, where the teacher always performs inferior to the student. Too frequent EMA will cause the teacher model soon collapsed to the student model, which causes forgetting problem and limits the following learning. Lowering the EMA frequency to every epoch (EMA\_epoch1) or every 5 epoch (EMA\_epoch5) performs better, which satisfies our theoretical analysis to keep total updating steps $n$ properly small. Our KC-EMA which empirically performs EMA every 5 epoch with adaptive momentum attains the best result.

\section{Conclusion and future work}
This paper propose a new setting for instance incremental learning task where no old data is available and the target is to enhance the base model with only new observations each time. The new IIL setting is more practical in real deployment either in regards of fast and low-cost model updating or data privacy policy. To tackle the proposed problem, a new decision boundary-aware distillation with knowledge consolidation method is presented. Benchmarks based on existing public datasets are established to evaluate the performance. Extensive experiments demonstrate the effectiveness of the proposed method. However, gap between the IIL model and full-data model still exists. In IIL, the future work can be done in following directions: 1) Narrowing the gap between IIL model and full-data model; 2) Few-shot IIL; 3) A better manner to accumulate knowledge than the proposed KC.
{
    \small
    \bibliographystyle{ieeenat_fullname}
    \bibliography{main}
}

\clearpage
\setcounter{page}{1}
\maketitlesupplementary

\section{Details of the theoretical analysis on KC-EMA mechanism in IIL}
\label{sec:limitation}
In Sec. 4.3 of the manuscript, we theoretically analyze the feasibility of applying a model EMA-like mechanism in the IIL. Here, gives more derivation details of the Eq. 7.

Given the relationship between the teacher model and the student model as \cref{eq:EMA},

\begin{equation}\label{eq:EMA}
    \theta_t^n = \alpha \theta_t^{n-1} + (1-\alpha) \theta_s^n
\end{equation}

\noindent where $\theta_t^n$ are the parameters of the teacher model and $\theta_s^n$ are the parameters of the student model. $\alpha \in [0, 1]$ is the updating momentum. 

After $n$ step's updating, it's not hard to get the parameters of the teacher model as \cref{eq:tweight}.
\begin{equation}
    \begin{aligned}\label{eq:tweight}
    \theta_t^n &= \alpha \theta_t^{n-1} + (1-\alpha)\theta_s^n \\
    &= \alpha^n \theta_t^0 + \alpha^{n-1}(1-\alpha)\theta_s^1 + \cdots \\ 
    & ~~~~~~~~+ \alpha^i(1-\alpha)\theta_s^{n-i} + \cdots + (1-\alpha)\theta_s^n
    \end{aligned}
\end{equation}

Assuming the model $M_0$ in base phase and the student model $M_t^s$ in the current phase are well trained on base data and new data, respectively, we have the derivative of the old task $\mathcal{L}_{old}$ on teacher model and the derivative of the new task $\mathcal{L}_{new}$ on student model as \cref{eq:derivatives}.

\begin{equation}\label{eq:derivatives}
    \frac{d\mathcal{L}_{old}}{d\theta_t^0}\approx0, \frac{d\mathcal{L}_{new}}{d\theta_s^n}\approx0
\end{equation}

The derivative of the old task(s) and the new task on the teacher model in current IIL phase is 

\begin{equation}
    \begin{aligned}
    \frac{d(\mathcal{L}_{old}+\mathcal{L}_{new})}{d\theta_t^n} &= \frac{d\mathcal{L}_{old}}{d\theta_t^n} + \frac{d\mathcal{L}_{new}}{d\theta_t^n}\\ 
    &= \frac{d\mathcal{L}_{old}}{d\theta_t^0}\frac{d\theta_t^0}{d\theta_t^n} + \frac{d\mathcal{L}_{new}}{d\theta_s^n}\frac{d\theta_s^n}{d\theta_t^n} \\
    &=\frac{1}{\alpha^n} \cdot \frac{d\mathcal{L}_{old}}{d\theta_t^0} + \frac{1}{1-\alpha}\cdot \frac{d\mathcal{L}_{new}}{d\theta_s^n}
    \end{aligned}\label{eq:td}
\end{equation}

Therefore, we get Eq. (7) and many conclusions can be drawn based on it as we present in the manuscript. Notably, as we assume the student is fully trained on new data, we set a freezing period during which we only train the student without implementing KC-EMA. In the manuscript, we empirically set the freezing period to 10 epochs.

\textbf{Limitation.} Our method may accumulate errors after a long consecutive IIL tasks. For example, in the $i^{th}$ IIL task, the old model should consider the base task and the previous $i-1$ IIL task. The old task's derivative on the parameters of teacher becomes 

\begin{equation}
    \frac{d\mathcal{L}_{old}}{d\theta_t^i}> \frac{d\mathcal{L}_{base}}{d\theta_t^0}+ \frac{d\mathcal{L}_{1}}{d\theta_t^1} + \cdots + \frac{d\mathcal{L}_{i-1}}{d\theta_t^{i-1}}
\end{equation}

As $0\approx \frac{d\mathcal{L}_{base}}{d\theta_t^0} < \frac{d\mathcal{L}_{1}}{d\theta_t^1} < \cdots < \frac{d\mathcal{L}_{i-1}}{d\theta_t^{i-1}}$, we can no longer deem $\frac{d\mathcal{L}_{old}}{d\theta_t} \approx 0$ after a sufficiently long sequence of IIL tasks. Therefore, the performance of teacher model will degrade along with the increasing tasks. However, such a kind of error accumulation is slow and will not affect the new knowledge learning. \textit{Error accumulation in task-level is a common problem in all incremental algorithms since no access to the old data.} In our experiments, we didn't observe significant performance degradation. One of the reason is that not all new samples are different with the old data. In another word, the old knowledge is possible to be trained on some new samples in instance-incremental learning. But for real application, we still suggest that the model should be rectified by a full training after a serials of incremental learning. 

\section{Algorithm overview}
The whole process of the proposed method is illustrated in Algorithm \ref{alg:method}. Codes will be released to public upon publication.

\begin{algorithm}[t]
    \caption{Overview of the proposed method}\label{alg:method}
    \textbf{Input:} new observations $D_n(t)=\left\{(x_i, y_i): y_i\in C\right\}_{i=1}^N$
    \textbf{Output:} model $M_t$
    \begin{algorithmic}[1]    
    \Require a trained model $M_{t-1}$    
    \State \textit{\# Boundary distillation}
    \State Initialize the teacher model  $T \leftarrow M_{t-1}$ and student model $S \leftarrow M_{t-1}$
    \For {epoch $e$ in $E$}
    \For {batch $B$ in $\lfloor N/B\rfloor$}
    \State \textit{\# Dusting the input space} 
    \State      $x^{'} \leftarrow norm(x) + G(\mu, \delta)$, where $x\in B$
    \State obtain predictions $p_t(x) \leftarrow T(x)$, $p_s(x) \leftarrow S(x)$, $p_t(x^{'}) \leftarrow T(x^{'})$, $p_s(x^{'}) \leftarrow S(x^{'})$    
    \State calculating fused label $y_f(x)$ according to \cref{eq:smooth_l}
    \State calculating $\mathcal{L}_{cls}$ according to  \cref{eq:train}
    \State updating student model S by optimizing $\mathcal{L}_{cls}$
    \EndFor
    \State \textit{\# Knowledge consolidation:}
    \If {$e>e_0\; and \;e\%5=0$}
    \State $\alpha \leftarrow \min(\alpha_0,1-\frac{e}{e+e_w}) $
    \State $\theta_t \leftarrow \alpha \theta_t + (1-\alpha) \theta_s$
    \EndIf
    \EndFor
    \State Return $M_t \leftarrow T$
    \end{algorithmic}
\end{algorithm}

\section{Dataset details}
\label{sec:dataset}
Corresponding to the definition of our new incremental-instance learning (IIL) task, we reorganize the public dataset Cifar-100~\cite{krizhevsky2009learning} and ImageNet-100~\cite{russakovsky2015imagenet} to establish the benchmark in this research topic. 

\textbf{Cifar-100}~\cite{krizhevsky2009learning} is the most commonly used dataset in incremental learning. It contains 60K $32\times32$ images of 100 classes. Each class has 500 images for training and 100 images for testing. In class-incremental learning, the training set is further split to many small subsets in terms of class. In the IIL, we randomly choose 25K images as the base dataset $D(0)$ to train a base model $M_0$. The remaining 25K images are randomly divided into 10 incremental sets. Thus, $D_n(t)=\frac{1}{10}D(0)$. Each incremental set contains samples from all classes but might has different number. As a result, long-tailed problem could exist in the incremental training, which is more realistic as real scenarios. 

\begin{figure}[t]
    \centering
    \includegraphics[width=\linewidth]{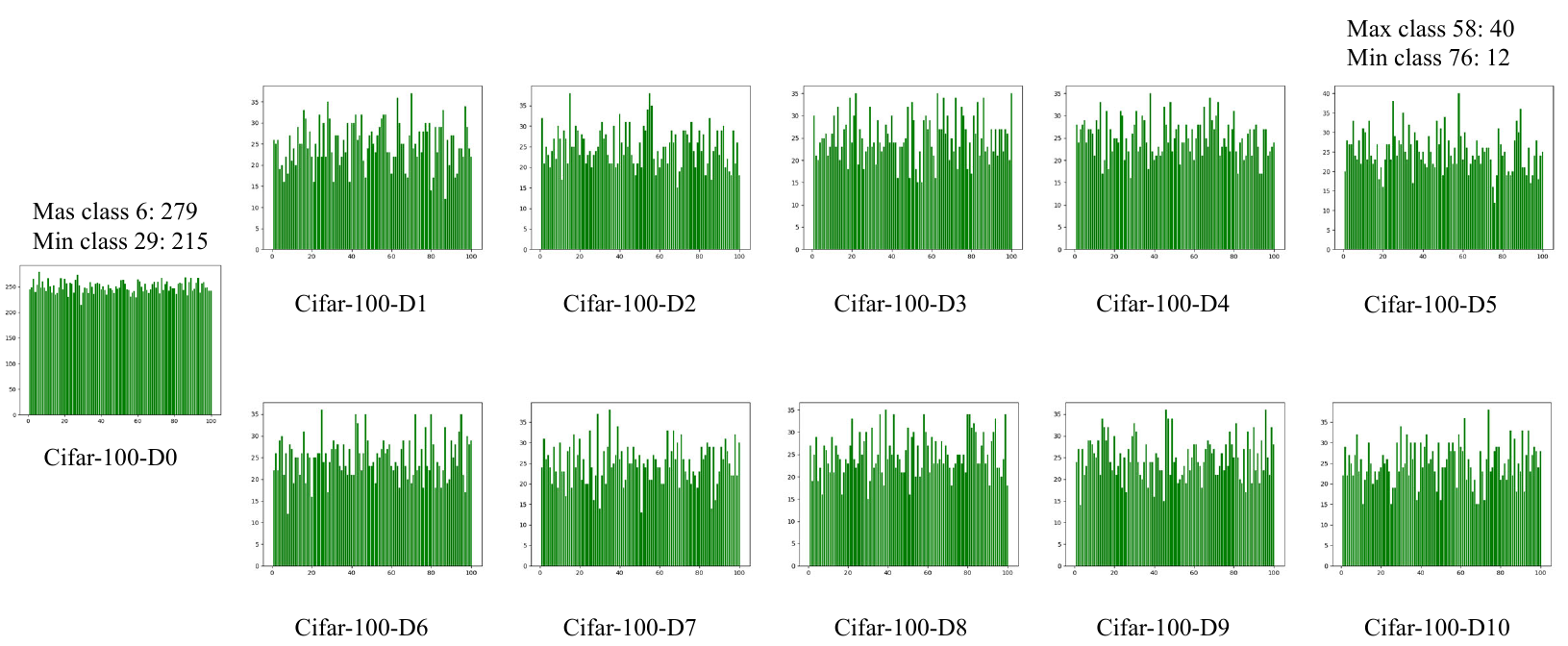}
    \caption{Number of images in each class in the base set and incremental sets of Cifar-100. ``D0" means the base set and ``Di" denotes the incremental set in task i.}
    \label{fig:dis_cifar}
\end{figure}

The distributions of image number contained in each class in the base set and incremental sets are shown in~\cref{fig:dis_cifar}. As can be seen, the base dataset has a more balance sample distribution, where the largest class is class 6 with 279 images. The smallest class is class 29 and has 215 images. Incremental dataset ``D5" has the most imbalance number of samples between classes, where the maximum number is 40 in class 58 and the minimum number is 12 in class 76. That is, the highest imbalance ratio is $3.33:1$.

\begin{figure}[t]
    \centering
    \includegraphics[width=\linewidth]{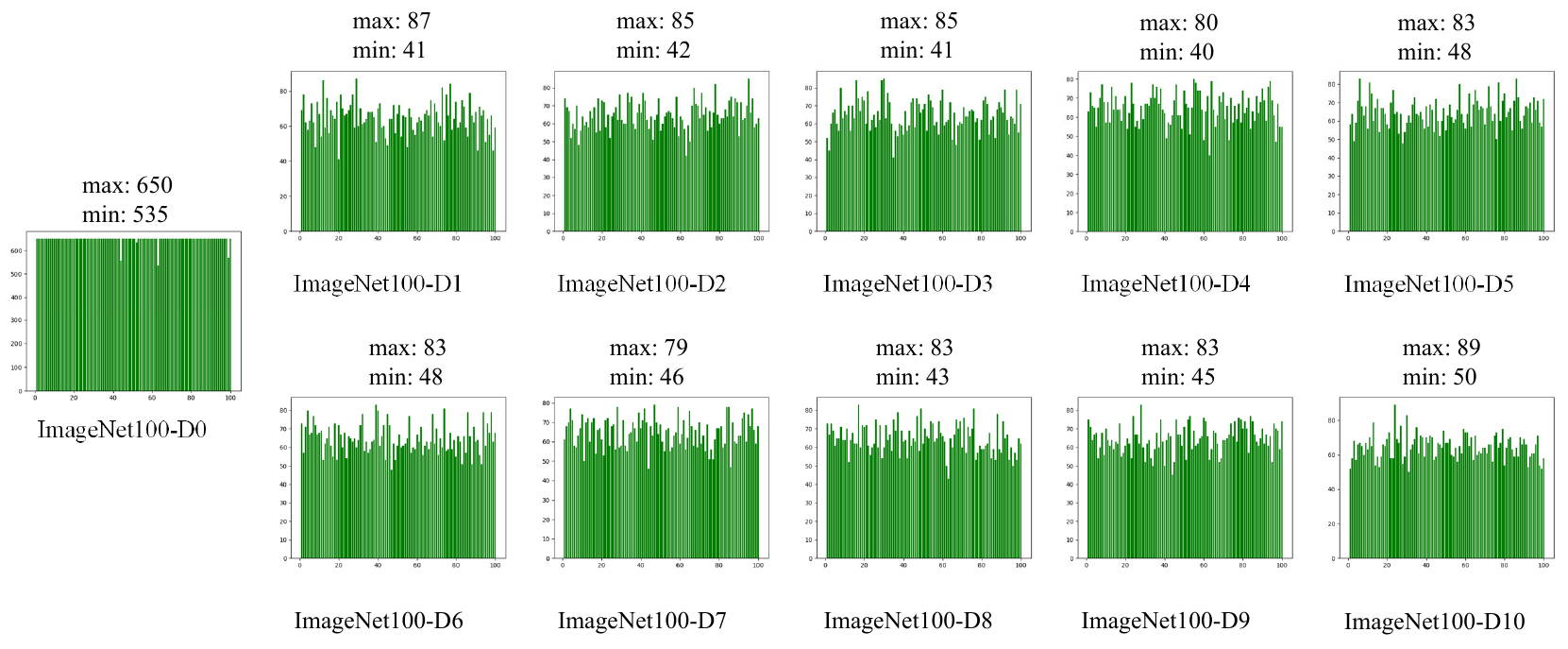}
    \caption{Number of images in each class in the base set and incremental sets of ImageNet-100. ``D0" means the base set and ``Di" denotes the incremental set in task i. ``max" and ``min" shows the most and least samples contained in a class, respectively.}
    \label{fig:dis_imagenet}
\end{figure}

\textbf{ImageNet-100}~\cite{russakovsky2015imagenet} is a subset of the ImageNet-1000 which consists of 1.2 million training images and 150K testing images from 1000 classes. Following Douillard et al.~\cite{douillard2020podnet}, we randomly select 100 classes and split it into 1 base set with half of the training images and 10 incremental sets as Cifar-100. In detail, the base set contains Test set is utilized as the same. The incremental sets in ImageNet-100 are more balance between classes ($max(\frac{n_{max}}{n_{min}})= 2.12:1$). The base dataset contains 64696 images and each incremental dataset contains 6470 images. The test set has 5000 images in total.

\section{Implementation details}
\textbf{Details of our method.} Following previous works, we adopt the standard resnet-18 as our backbone network in all experiments on Cifar-100 and ImageNet. All results are attained by training 60 epochs as iCarl~\cite{rebuffi2017icarl} did. Initial learning rates for base model are 0.001 on Cifar-100 using Tensorflow and 0.1 on ImageNet using Pytorch. Our experiments are first implemented based on Tensorflow to compared with iCarl on Cifar-100. While some other methods are implemented with Pytorch on ImageNet. To avoid the reproduction error, we also implement our method using Pytorch on ImageNet. For incremental learning, the learning rate is decayed with 0.1 compare to base model training. $\lambda$ in \cref{eq:train} is set as 0.1. $\alpha_0$ is 0.99 and $e_w$ is 500 in \cref{eq:momentum}.

\textbf{Selection of compared methods.}
As few existing method is proposed for the IIL setting, we reproduce several classic and SOTA CIL methods by referring to their original code or paper with the minimum revision. \\
\textbf{LwF}~\cite{li2017learning} is one of the earliest incremental learning algorithms based on deep learning, which propose to use knowledge distillation for resisting knowledge forgetting. Considering the significance of this method, a lot of CIL methods are still comparing with this baseline. \\
\textbf{iCarl}~\cite{rebuffi2017icarl} is base on the LwF method and propose to use old exemplars for label-level knowledge distillation. \\
\textbf{PODNet}~\cite{douillard2020podnet} implements old knowledge distillation at the feature level, which is different from the former two. \\
\textbf{Der}~\cite{yan2021dynamically} which expends the network dynamically and attains the best CIL results given task id. Expending the neural network shows great power in learning new knowledge and retain old knowledge. Although in new IIL setting the adding of new parameters is limited, we are glad to know the performance of the method with dynamic network in the new IIL setting. \\
\textbf{OnPro}~\cite{wei2023online} uses online prototypes to enhance the existing boundaries with only the data visible at the current time step, which satisfies our setting without old data. Their motion to make the learned feature more generalizable to new tasks is also consist with our method to promote the model continually utilizing only new data.\\
\textbf{online learning}~\cite{he2020incremental} can be applied to the hybrid-incremental learning. Thus, it can be implemented directly in the new IIL setting. It proposes a modified cross-distillation by smoothing student predictions with teacher predictions for old knowledge retaining, which is different with our method to alter the learning target by fusing annotated label with the teacher predictions.

Besides above CIL methods, \textbf{ISL}~\cite{liang2022balancing} is one of the scarce IIL methods that can be directly implemented in our IIL setting. Different from our setting that aims to address all newly achieved instances, ISL is proposed for incremental sub-population learning. Hence, we not only applied ISL in our setting for comparative purposes but also evaluated ours following their setting.

Our extensive experiments real that neither existing CIL methods nor IIL methods can tame the proposed IIL learning problem. Existing methods, especially the CIL methods, primarily concentrate on mitigating catastrophic forgetting, demonstrating limited effectiveness in learning from new data. In real-world applications, enhancing the model with additional instances and achieving more generalizable features is crucial. This underscores the importance and relevance of our proposed IIL setting.

\textbf{Details for reproduction of compared methods.} To reproduce existing rehearsal-based methods, including iCarl~\cite{rebuffi2017icarl}, PODNet~\cite{douillard2020podnet}, Der~\cite{yan2021dynamically}, OnPro~\cite{wei2023online}, although no old data is available in the new IIL setting, we still set a memory of 20 exemplars per class. For all compared methods, we trained the base model for their own if necessary. For example, some of the methods utilize more comprehensive data augmentation methods and have a higher base performance than others. We kept the data augmentation part in reproducing these methods, even in the IIL learning phase.

For LwF~\cite{li2017learning}, as there is no need to add new classification head, we directly implement their training loss on the old classification head. iCarl~\cite{rebuffi2017icarl} is implemented using their provided codes on Tensorflow. PODNet~\cite{douillard2020podnet} and Der~\cite{yan2021dynamically} are implemented based on their own codes using Pytorch. For online learning~\cite{he2020incremental}, we reproduce it only using the part that is related to learning from new instances of old classes in the first training phase. ISL~\cite{liang2022balancing} is also implemented based on its original codes with the original setting of the learning rate, \ie lr=0.05 for the base training and lr=0.005 for the IIL learning phase. OnPro~\cite{wei2023online} utilize a quite strong data augmentation during training, which leads to a slightly low accuracy of the based model compared to other methods. As OnPro trains the base model for only one epoch in the online training, which is not sufficient to achieve a strong base model, we change it to 60 epochs for training the base model as other methods. For the IIL learning phase, we keep the one epoch setting as it is because we found training more epochs don't lead to better performance. The learning rate of OnPro used in training the based model is 5e-4 and decays with a ratio of 0.1 at epoch 12, 30, 50. In the IIL traing phase, lr=5e-7, which is far smaller than ours (0.01). 

\section{Visualization of dusted input images}
\begin{figure}[t]
    \centering
    \includegraphics[width=\linewidth]{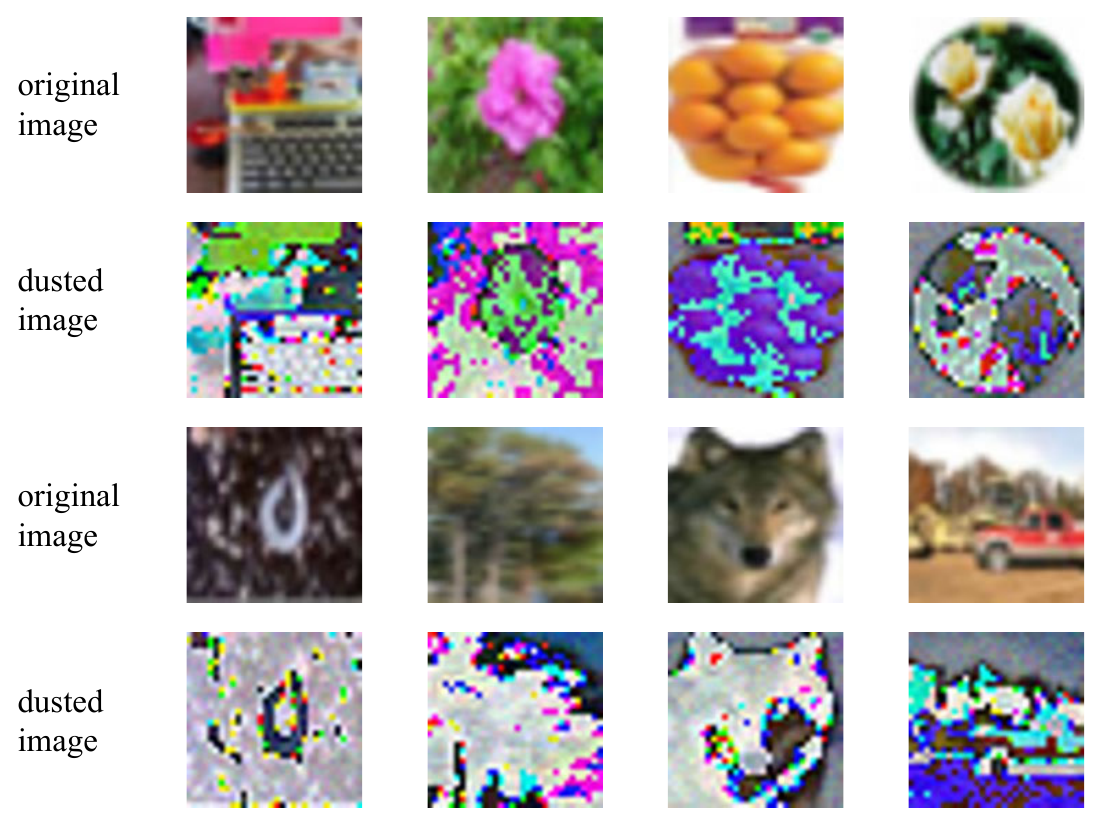}
    \caption{The original images and corresponding dusted images.}
    \label{fig:dustedimgs}
\end{figure}

To distill the decision boundary in an existing model, we proposed a module to dusted the input space with random Gaussian noise. By dusting the input space, we hope some samples can be relocated to the peripheral area of the learned decision boundary. Therefore, the intractable decision boundary can be manifested to some extent and distilled to the student model for knowledge retaining. The input space pollution is different with the image augmentation in the train process because of the large deviation and allowance of the polluted images to be classified to different classes besides their original labels. In fact, we hope the polluted images are prone to be classified to other category than the original category. The boundary can only be known when we know what is and what is not. The dusted input images is visualized in \cref{fig:dustedimgs}. It can be seen that the category of each image becomes vague after dusting.

\section{More experimental results}

\subsection{Ablation study on network size}
To investigate the impact of network size on the proposed method, we compare the performance of ResNet-18, ResNet-34 and ResNet-50 on the ImageNet-100. As shown in \cref{tab:imp_size}, the proposed method performs well with bigger network size. When the network size is larger, more parameters can be utilized for new knowledge learning with the proposed decision boundary-aware distillation. Hence, consolidating knowledge from student to teacher causes less forgetting.

\begin{table}
    \centering
    \resizebox{\linewidth}{!}{
    \begin{tabular}{c|c|c|c|c}
    \toprule
 & \multicolumn{2}{c|}{Cifar-100}& \multicolumn{2}{c}{ImageNet-100}\\
    \hline
         Backbone&  $PP$(\%)&  $\mathcal{F}$(\%)&  $PP$ (\%)& $\mathcal{F}$ (\%)\\
         \hline
         ResNet-18&  +4.93$_{\pm 0.21}$&  -1.86$_{\pm 0.08}$ & +2.33$_{\pm 0.13}$ & -5.97$_{\pm 0.13}$\\
         \hline
         ResNet-34&  +6.65$_{\pm 0.12}$&  -1.39$_{\pm 0.05}$ & +3.01$_{\pm 0.18}$ & -4.90$_{\pm 0.08}$\\
         \hline
         ResNet-50& +6.92$_{\pm 0.11}$& -1.51$_{\pm 0.05}$ & +1.68$_{\pm 0.07}$ &-4.52$_{\pm 0.98}$\\
    \bottomrule
    \end{tabular}}
    \caption{Impact of the network size on the proposed method.}
    \label{tab:imp_size}
\end{table}

\subsection{Ablation study on the task number}
As mentioned in Sec.~\ref{sec:limitation}, our method accumulates the error along with the consecutive IIL tasks. However, such a kind of error accumulates slowly and mainly affects the performance on old tasks, \ie forgetting rate. We further study the impact of task length on the performance of the proposed method by splitting the incremental data into different number of subsets. As shown in \cref{tab:task_length}, with the incremental of task number, the performance promotion changes less but the forgetting rate increased slightly. Minor variation of performance promotion reveals that the proposed method is stable in learning new knowledge, irrespective of the number of tasks. The acquisition of new knowledge primarily hinges on the volume of new data involved. Although we increase the task number in the experiments, the total number of new data utilized in IIL phase is the same. While increasing the task number will increase the EMA steps, which causes more forgetting on the old data. Experimental results in \cref{tab:task_length} well validate our analysis in Sec.~\ref{sec:limitation}.

Compared to the performance promotion, forgetting on the old data is negligible. Noteworthy, when the task number is relatively small, such as 5 in \cref{tab:task_length}, the proposed method slightly boosts the model's performance on the base data. This behavior is similar with full-data model, which demonstrates the capability of our method in accumulating knowledge from new data.
\begin{table}
    \centering
    \resizebox{\linewidth}{!}{
    \begin{tabular}{c|cccc}
    \toprule
         \textbf{Cifar-100}  & 5-task & 10-task & 15-task & 20-task \\
         \hline
        $PP (\%)$ & +4.92$_{\pm 0.14}$ & +4.93$_{\pm0.21}$  & +5.09$_{\pm0.08}$ & +4.99$_{\pm0.19}$\\
    \hline
        $\mathcal{F} (\%)$ & +0.61$_{\pm 0.02}$ & -1.86$_{\pm0.08}$ & -1.88$_{\pm0.09}$ & -2.35$_{\pm0.14}$ \\
    \bottomrule
    \end{tabular}}
    \caption{Performance of the proposed method with different IIL task numbers.}
    \label{tab:task_length}
\end{table}

\subsection{Detailed comparison between the KC-EMA and vanilla EMA}
\begin{figure}[t]
    \centering
    \includegraphics[width=\linewidth]{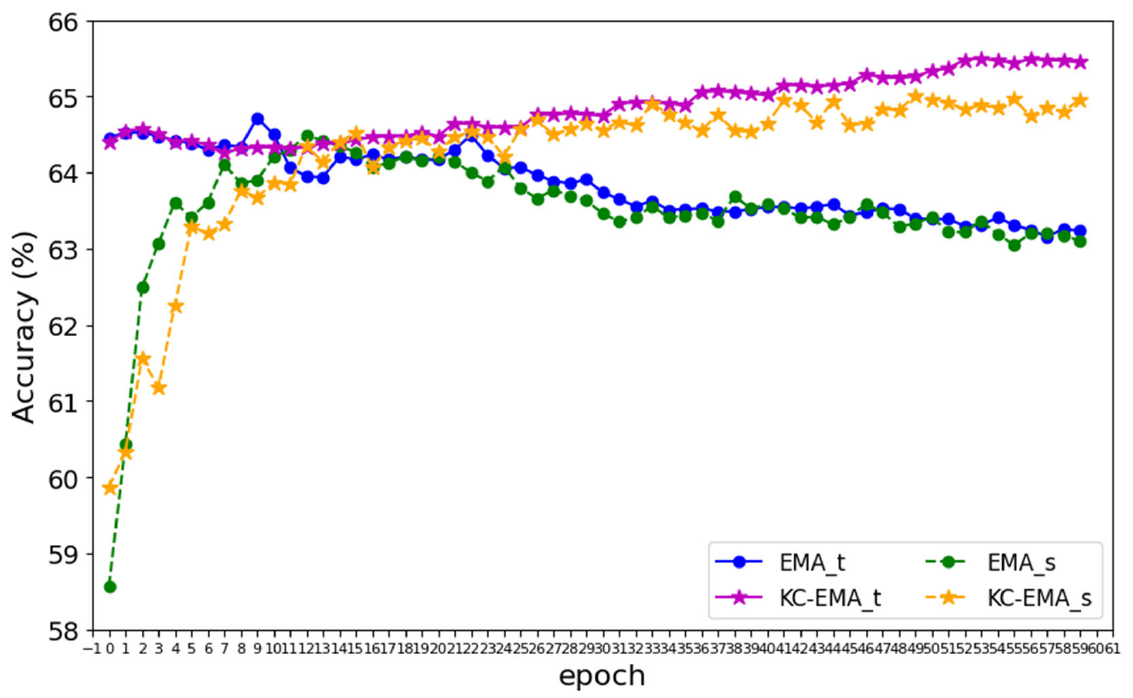}
    \caption{Comparison between the proposed KC-EMA and vanilla EMA during training in the first IIL task, where t denotes the teacher model and s denotes the student model. Result is achieved on Cifar-100.}
    \label{fig:compa_e1}
\end{figure}

The performance of vanilla EMA and the proposed KC-EMA during training is shown in \cref{fig:compa_e1}. As can be seen, the student model's accuracy initially plummets due to the introduction of new data. However, around the 10th epoch, there's a resurgence in accuracy for both the KC-EMA and vanilla EMA models. Therefore, we empirically set a freezing epoch of 10 in the proposed method. 

When EMA is applied post the 10th epoch, the teacher model in the vanilla EMA is rapidly drawn towards the student model. This homogenization, however, doesn't enhance either model. Instead, it leads to a decline in test accuracy due to overfitting to the new data. In contrast, with KC-EMA, both the teacher and student models exhibit gradual growth,, which indicates a knowledge accumulation in these two models. On one hand, consolidating new knowledge to the teacher model  improves its test performance. On the other hand, a teacher model equipped with new knowledge liberates the student model to learn new data. That is, constraints from the teacher in distillation is alleviated.

\end{document}